%% file: main.tex
\documentclass[10pt,twocolumn,letterpaper]{article}

\usepackage{cvpr}              

\input{preamble}

\definecolor{cvprblue}{rgb}{0.21,0.49,0.74}
\usepackage[pagebackref,breaklinks,colorlinks,citecolor=cvprblue]{hyperref}
\usepackage{multirow}
\usepackage[table,xcdraw]{xcolor}
\definecolor{mycolor1}{HTML}{5C3A9A}
\definecolor{mycolor2}{HTML}{6B913B}
\definecolor{purple}{RGB}{0,0,0}
\newcommand{\supp}[1]{\textcolor{magenta}{#1}}
\usepackage{booktabs}
\usepackage{graphicx}

\def\paperID{*****} 
\def\confName{CVPR}
\def\confYear{2024}

\title{From Pen to Pixel: Translating Hand-Drawn Plots into Graphical APIs via a Novel Benchmark and Efficient Adapter}

\author{
Zhenghao Xu$^{1}$, Mengning Yang$^{1,*}$\\
$^{1}$School of Big Data \& Software Engineering, Chongqing University, Chongqing, China.\\
{\tt\small hzx1290681496@gmail.com, mnyang@cqu.edu.cn}\\
{\small $^{*}$ Corresponding author}
}

\begin{document}
\maketitle
\begin{abstract}
As plots play a critical role in modern data visualization and analysis, Plot2API is launched to help non-experts and beginners create their desired plots by directly recommending graphical APIs from reference plot images by neural networks.
However, previous works on Plot2API have primarily focused on the recommendation for standard plot images, while overlooking the hand-drawn plot images that are more accessible to non-experts and beginners.
To make matters worse, both Plot2API models trained on standard plot images and powerful multi-modal large language models struggle to effectively recommend APIs for hand-drawn plot images due to the domain gap and lack of expertise.
To facilitate non-experts and beginners, we introduce a hand-drawn plot dataset named HDpy-13 to improve the performance of graphical API recommendations for hand-drawn plot images.
Additionally, to alleviate the considerable strain of parameter growth and computational resource costs arising from multi-domain and multi-language challenges in Plot2API, we propose Plot-Adapter that allows for the training and storage of separate adapters rather than requiring an entire model for each language and domain.
In particular, Plot-Adapter incorporates a lightweight CNN block to improve the ability to capture local features and implements projection matrix sharing to reduce the number of fine-tuning parameters further.
Experimental results demonstrate both the effectiveness of HDpy-13 and the efficiency of Plot-Adapter.
\textbf{The code is available in \supp{https://github.com/xuzhengh/Hand-Drawn-Plot2API}.}
\end{abstract}
\section{Introduction}
\label{sec:intro}
\begin{figure}[t]
  \centering
   \includegraphics[width=0.9\linewidth]{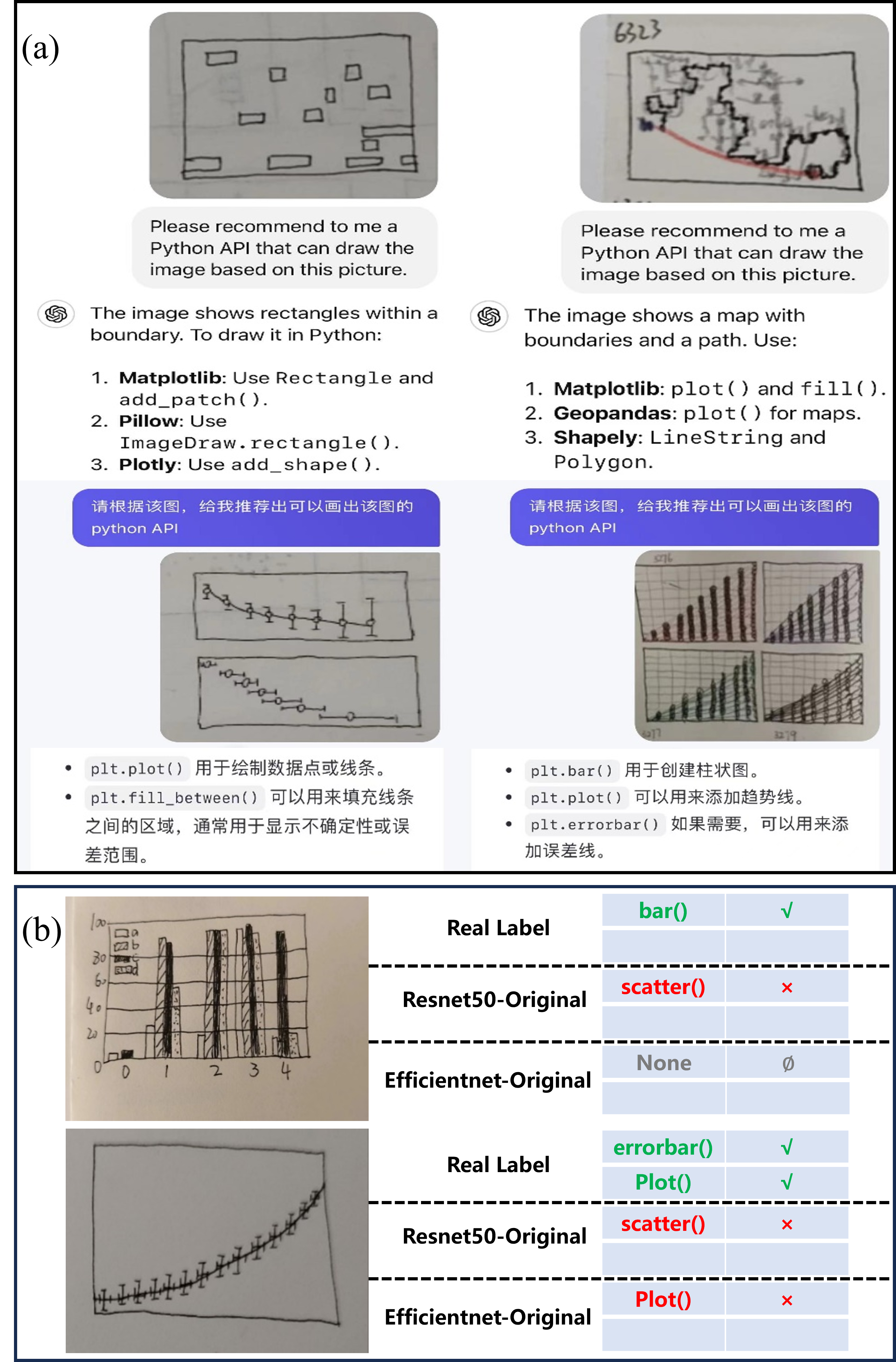}
   \caption{(a) Example of API recommendations by large language models. (b) API recommendation results based on hand-drawn plot images using ResNet-50 and EfficientNet models trained on a standard plot dataset.}
   \label{fig:one}
\end{figure}
Data visualization plays a vital role in scientific research, business, and engineering~\cite{vision-important1,vision-important2}. 
As an intuitive representation of data, plots not only improve the efficiency and accuracy of information transmission but also provide direct support for data analysis and decision-making. 
Therefore, plotting functions have been widely integrated into programming languages through graphical application programming interfaces (APIs). 
These graphical APIs offer numerous plotting functions and tools, enabling developers and programmers to create intuitive plots~\cite{interaction}, and have become essential components of the core libraries in programming languages~\cite{matplotlib, ggplot2}.

However, the diversity and similarity of plots make it challenging for developers, especially for non-experts and beginners, to find the correct APIs, which often leads to significant time costs on search and validation~\cite{hard-to-find-api}. 
To address this issue, Wang et al.~\cite{plot2api} first proposed Plot2API, where plot images are input into neural networks to output the target APIs in a plot-to-API manner directly. 
Specifically, Plot2API transforms the graphical API recommendation into a multi-label image classification task~\cite{multi-Label-Learning-Algorithms} and devises a semantic parsing guided neural network (SPGNN) to recommend APIs through semantic features. 
Furthermore, observing the data imbalance issue in Plot2API, Qin et al.~\cite{mstil} proposed cross-language shape-aware plot transfer learning (CLSAPTL) and cross-language API semantic similarity-based data augmentation (CLASSDA) to alleviate the imbalance between different APIs.
Besides, a balanced multi-label softmax loss (BMLSL) is introduced to address the positive-negative imbalance.
\begin{figure}[t]
  \centering
   \includegraphics[width=0.9\linewidth]{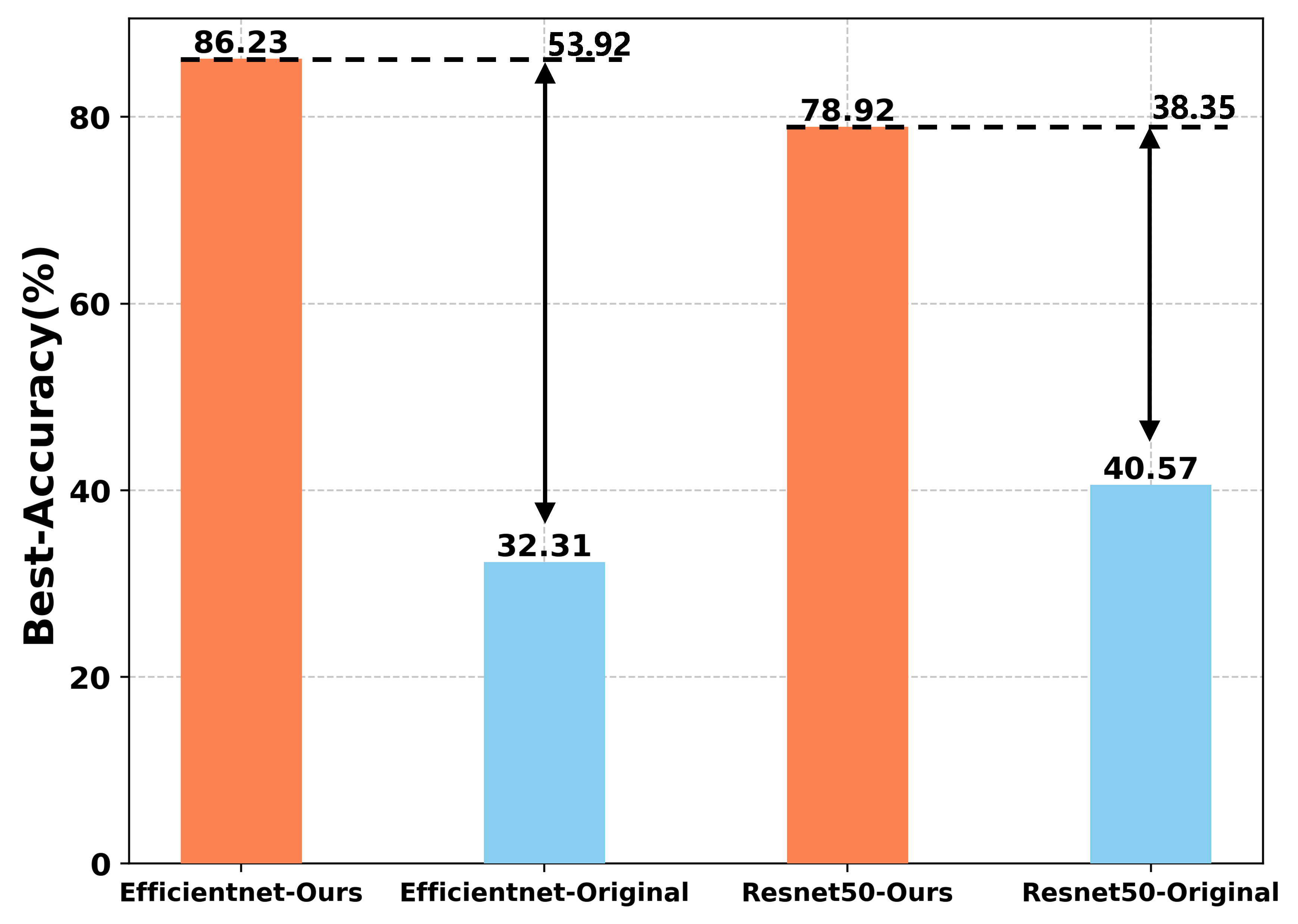}
   \caption{Accuracy comparison in the task of API recommendation based on hand-drawn plot images: orange denotes models trained on our dataset, blue denotes models trained on a standard plot dataset.}
   \label{fig:two}
\end{figure}

Previous works in Plot2API have primarily focused on standard plot images, with limited attention to hand-drawn plot images~\cite{plot2api,mstil}. 
However, it is usually tricky for non-experts and beginners to find standard plot images, including their target visualization contents as references, especially for some unusual APIs.
In such cases, hand-drawn plot images become a more straightforward and quicker alternative, highlighting the importance of developing corresponding Plot2API methods. 
Fig.~\ref{fig:one} (b) and Fig.~\ref{fig:two} illustrate the unsatisfactory recommendation performance of models trained with standard plot images when applied to hand-drawn ones, which could be attributed to the domain gap between different types of plot images. 
Besides, as shown in Figure~\ref{fig:one}(a), multi-modal large language models (e.g., ChatGPT, Tongyi Qianwen) \cite{largemodel1} also fall short in recommending APIs for hand-drawn plot images. 
Due to a lack of related expertise~\cite{largemodel, largemodel2}, their recommendations often fail to meet developers' specific needs. 
These phenomena further demonstrate the necessity of developing targeted Plot2API models for hand-drawn plot images.

To facilitate non-experts and beginners, we construct a hand-drawn plot dataset called HDpy-13 to help researchers train Plot2API models for hand-drawn plot images. 
Firstly, we take the public Python-13 dataset~\cite{plot2api} as a reference and gather volunteers to redraw the samples and save them as images manually. 
Python-13 contains 13 Python APIs and corresponding 6350 standard plot images originating from posts in the Stack Overflow community, which meet the needs of most developers.
The volunteers not only hand-draw most images of Python-13 but also provide additional samples for rare APIs while reducing that of the common ones. 
Finally, HDpy-13 consists of 5163 plot images. 
%
%
We train ResNet-50 \cite{resnet} and EfficientNet \cite{efficientnet} models separately on the train-set of Python-13 and HDpy-13 while evaluating both on the test-set of HDpy-13. 
The results are reported in Fig.~\ref{fig:two}.
The results demonstrate that the ResNet and EfficientNet models trained on HDpy-13 (Ours) achieve accuracies of 70\% and 80\%, respectively, significantly outperforming the same models trained on the standard plot dataset (Original), which only achieve accuracies of 30\% and 40\%.

In addition to the API recommendation issue on hand-drawn plot images, we have also observed a unique challenge that was overlooked in previous work.
Specifically, the necessity of training and storing separate models for different programming languages and domains results in significant parameter growth and computational resource costs~\cite{qin2025no}.
To address this issue, we propose a novel Plot-Adapter to train and store only separate adapters instead of the entire model for each language and domain while maintaining competitive recommendation performance.
To achieve the latter, we replace the simple nonlinear transformation in the vanilla adapter with a lightweight CNN block to enhance local spatial feature capture capacity while introducing a projection matrix sharing strategy to counteract the additional fine-tuning parameters from the CNN block.
Extensive experiments on datasets from different programming languages and domains demonstrate both the effectiveness and efficiency of our proposed Plot-Adapter.

Our contributions are three-fold: 
1)We observed a lack of support for hand-drawn image recommendations in Plot2API. 
Therefore, we propose HDpy-13 to enhance the API recommendation capabilities of Plot2API for hand-drawn plot images.
2)We devise Plot-Adapter to alleviate parameter growth and computational overhead in multi-domain, multi-language scenarios, while enhancing the backbone's ability to capture local spatial features.
3)We conduct experiments across diverse languages and domains to validate the effectiveness of HDpy-13 and the efficiency of Plot-Adapter.

\section{Related work}
\label{sec:related}
\subsection{API Recommendation and Plot2API}
%
Numerous well-established approaches primarily utilize natural language processing (NLP) techniques to generate API recommendations from textual queries and source code~\cite{NLP1, NLP2, NLP3,pyart, rack}. 
For example, Thung et al.~\cite{automatic2013} developed an automatic method by analyzing natural language in feature requests, helping developers translate requirements into APIs. 
Gu et al.~\cite{deepapi} proposed DeepAPI, a deep learning model with an RNN encoder-decoder to map text queries to API sequences. 
In addition, Nguyen et al.~\cite{text2api} introduced T2API, a statistical translation model that transforms natural language into API code templates. 
Nguyen et al.~\cite{focus} developed FOCUS, a context-aware collaborative filtering system that recommends APIs by mining usage patterns from open-source software repositories. 
Ling et al.~\cite{gnn} introduced a system based on graph neural networks (GNNs) that constructs API call graphs to identify relationships between APIs.
In contrast to traditional API recommendation tasks, the emerging Plot2API aims to exploit discriminative features of plot images to recommend corresponding graphical APIs directly in a plot-to-API manner. 
Plot2API can be regarded as a multi-label image classification challenge. 
Wang et al.~\cite{plot2api} first launched Plot2API and proposed the semantic parsing guided neural network (SPGNN) to extract semantic features from plot images for API recommendations. 
Nonetheless, serious overfitting issues are caused by multiple factors in Plot2API, which limit its wider application. 
Qin et al.~\cite{mstil} enhanced Plot2API using MSTIL, which incorporates cross-language shape-aware plot transfer learning (CLSAPTL) and cross-language API semantic similarity-based data augmentation (CLASSDA) to alleviate the imbalance between the number of samples of different APIs. 
Moreover, a balanced multi-label softmax loss (BMLSL) is introduced to address the imbalance between positive and negative samples. 
%
%
%
To facilitate developers, we propose the HDpy-13 dataset to boost the application of Plot2API on hand-drawn plot images.
\subsection{Adapter}
As an efficient parameter fine-tuning approach for the Transformer architecture, adapter technology has shown notable flexibility and efficiency in computer vision~\cite{adapter1,adapter2,adapter3,adapter4,clipadapter}. 
For instance, in dense prediction tasks, ViT-Adapter~\cite{vitadapter} employs spatial prior modules to enhance Vision Transformers (ViT), enabling efficient object detection and semantic segmentation.
In image generation, T2I-Adapter~\cite{t2iadapter} enhances control capabilities over text-to-image diffusion models, while ControlNet \cite{text2imgadapter} refines spatial conditioning to achieve precise manipulation of generated images.
Moreover, Bi-Adapter \cite{biadapter} and AdaptFormer \cite{adaptformer} also illustrate the adaptability of adapters in image, video, and image recognition tasks, achieving task adaptation and knowledge transfer with minimal parameter adjustments.
We propose Plot-Adapter to achieve programming language switching with minimal parameter adjustments while maintaining competitive recommendation performance.
\section{Dataset}
\label{sec:data}
\begin{figure}[t] 
  \centering
  \includegraphics[width=1.0\linewidth]{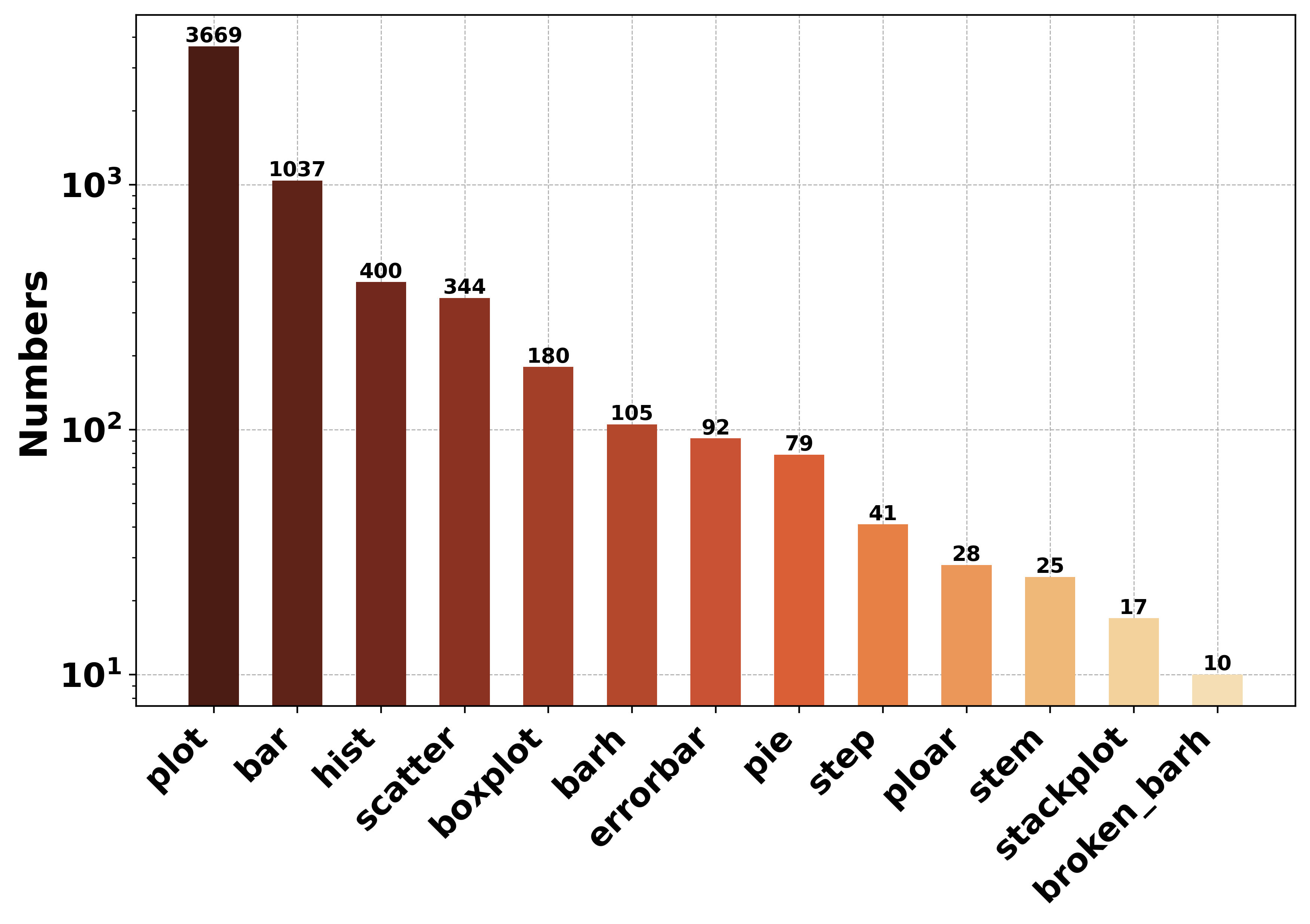}
  \caption{Dataset distribution of each API in HDpy-13.}
  \label{fig:four}
\end{figure}
For non-experts and beginners working with Plot2API~\cite{plot2api}, finding standard plot images that meet their visualization needs in order to obtain API recommendations can be challenging.
In this case, hand-drawn plot images become a simpler and quicker alternative.
However, previous Plot2API models trained on standard plot images fail to achieve satisfactory recommendation performance when applied to hand-drawn ones, which is mainly attributed to the well-known domain gap between these two types of reference images. 
To boost graphical API recommendations on hand-drawn plot images and satisfy the visualization requirements of users, we propose a hand-drawn plot dataset called HDpy-13.
%
%
%
\subsection{Dataset Summary}
The HDpy-13 dataset consists of 13 Python graphical APIs: \texttt{bar}, \texttt{barh}, \texttt{boxplot}, \texttt{broken barh}, \texttt{errorbar}, \texttt{hist}, \texttt{pie}, \texttt{plot}, \texttt{polar}, \texttt{scatter}, \texttt{stackplot}, \texttt{stem}, and \texttt{step}. 
It includes 5163 hand-drawn plot images, each with an average resolution of 1056×822 pixels. 
The dataset distribution is depicted in Figure~\ref{fig:four}. 
The training set consists of 4160 hand-drawn plot images, while the validation set consists of 1003, maintaining a ratio of approximately 4:1. 
All hand-drawn plot images in HDpy-13 have been annotated with relevant APIs in the Python programming language.

\begin{figure*}[t]
  \centering
  \includegraphics[width=1\linewidth]{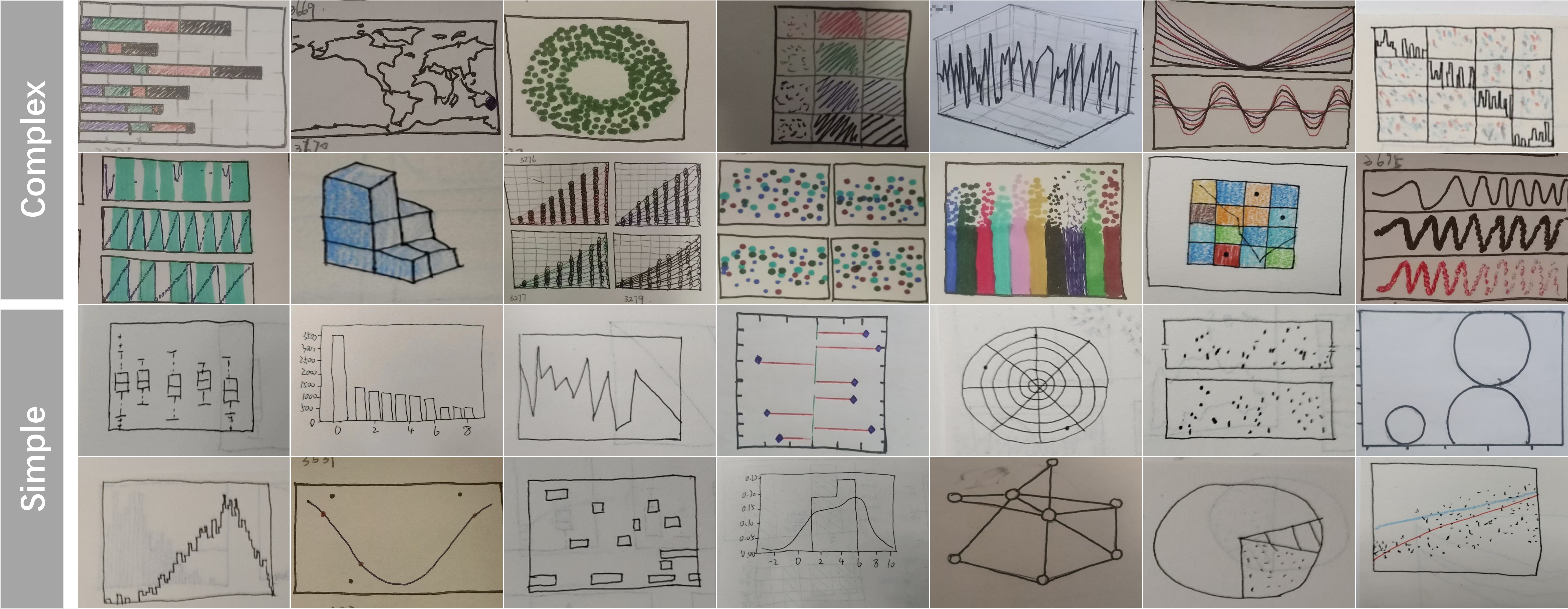}
  \caption{Sample hand-drawn plot images from HDpy-13 dataset showcasing various APIs and styles.}
  \label{fig:three}
\end{figure*}
\subsection{Data Collection, Annotation, Filtering}
HDpy-13 is primarily constructed with reference to the publicly available Python-13 dataset~\cite{plot2api}.
Python-13 is collected from posts in the Stack Overflow community, which provides authentic and widely applicable real-world samples. 
It contains over 6000 standard plot images with multi-label annotations corresponding to 13 Python graphical APIs. 
Therefore, we choose Python-13 as the reference for HDpy-13. 
We recruited 30 volunteers to manually redraw the samples from Python-13 and save them as images.

In addition to replicating some of the original plot images, we also introduce modifications to the hand-drawn plot images by changing colors, line styles, and overall layout. 
To diversify the hand-drawn styles and simulate the complexity of real-world application scenarios, we randomly assigned each volunteer the sample from Python-13. 
This prevents a single volunteer from producing all hand-drawn samples corresponding to a particular API, avoiding uniformity in the drawing styles. 
Notably, volunteers not only hand-draw most of the images from Python-13 but also create additional samples for rare APIs while reducing the number of samples for the common ones.
For the replicated samples from Python-13, we follow the original annotations from Python-13.
For the created plot images, volunteers manually annotated each hand-drawn plot image in detail, referencing the 13 graphical APIs in Python-13. 
Finally, we collected over 5000 candidate hand-drawn plot images to construct HDpy-13.

Specifically, we designed strict rules for selecting the candidate hand-drawn plot images. 
First, we remove the low-resolution hand-drawn samples, which have a resolution below the standard threshold (set at 300 x 300 pixels). 
Second, we invite volunteers to cross-evaluate the hand-drawn plot images, filtering out those with extremely rough details or poor drawing quality to avoid disturbing the final model training. 
Finally, the selected hand-drawn plot images are converted to PNG format, and corrupted files are removed.
%
%
The SSIM score between HDpy-13 and Python-13 is 0.594, indicating that while emulating essential elements of the original samples, HDpy-13 also maintains sufficient diversity.
\subsection{Dataset Characteristics}
We report the statistical data of the dataset in Figure~\ref{fig:four}. 
Common simple graphical APIs, such as \texttt{plot()}, tend to have a larger number of samples, while rare specialized APIs, such as \texttt{polar()}, usually have fewer ones. 
This indicates that common simple APIs are more frequently used due to their broad applicability and ease of use. In contrast, rare specialized APIs are used less often because of their limited application scope and smaller target user base.
Figure~\ref{fig:three} further shows the representative hand-drawn samples from HDpy-13. 
In real-world application scenarios, significant differences in the quality and resolution of hand-drawn plot images usually exist due to significant differences in users' skill levels and devices. 
To simulate this complexity, our dataset includes a variety of plot structures, ranging from simple to complex. 
HDpy-13 contains both samples drawn with a single API and those created through a combination of multiple APIs, which have varying hand-drawn fineness, from high to low. 
Unlike standard samples, in HDpy-13, simple plots generally exhibit higher fineness due to their clear lines and essential elements, while complex plots tend to have lower fineness. 
This reflects the real challenges of hand-drawn plot images. 
By including samples with varying fineness and complexity, our dataset effectively simulates real-world application scenarios, thus providing broad applicability. 
We suggest that our proposed HDpy-13 offers a valuable resource for research on both Plot2API and hand-drawn plot images.

\section{Methodology}
\label{sec:net}
\subsection{Preliminary}
\begin{figure*}[t]
  \centering
   \includegraphics[width=1\linewidth]{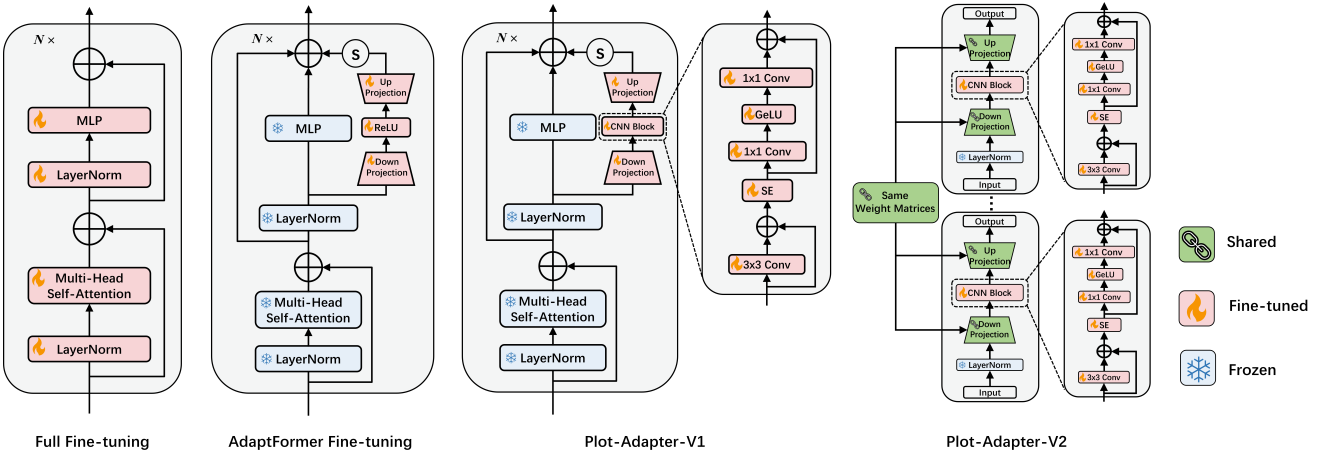}
   \caption{Comparison of previous fine-tuning strategies and our Plot-Adapter. Plot-Adapter-V1 enhances the backbone local feature capture ability by integrating token mixer, re-attention layer, and channel mixer. Plot-Adapter-V2 further improves parameter efficiency by introducing a projection matrix sharing strategy while preserving recommendation performance.}
   \label{fig:net5}
\end{figure*}

\textbf{Plot2API.} Plot2API is an API recommendation task first formally launched by \cite{plot2api}, which aims at recommending appropriate graphical APIs based on the specified programming language and the input image.
On the one hand, as the plots of each image could be drawn by multiple functions from a set of graphical APIs, Plot2API can be transformed as a multi-label image classification task.
On the other hand, each plot can also be represented through graphical API combinations from different programming languages.
Specifically, in Plot2API, given the $j$-th element of the programming language set $\mathcal{G}$, the graphical API set can be defined as \( A = \{a_i\}^{m}_{i=1} \), where \( m \) represents the corresponding number of APIs.
And for the input image \( \mathbf{I} \), the target of Plot2API is to output a binary vector \( \mathbf{Y}_j = \{ y_i\}^{m}_{i=1} \), where \( y_i = 1 \) if \( \mathbf{I} \) is associated with \( a_i \), and \( y_i = 0 \) otherwise.
Therefore, \textit{traditional} Plot2API could be denoted as learning a one-way mapping \( F(\cdot) \) from \( \mathbf{I} \) to \( \mathbf{Y}_j \):
\[
\mathbf{I} \overset{F_{\textcolor{red}{{\theta}}_j}(\cdot)}{\rightarrow} \mathbf{Y}_j, \tag{1}
\]
where \(\theta_j\) represents the parameters corresponding to \( F(\cdot) \) and the \textcolor{red}{\textbullet} color indicates \textcolor{red}{learnable}.\\
\textbf{Vision Transformer.} Vision Transformer (ViT) was first proposed by \cite{vanillavit} and rapidly became popular in the field of computer vision, which is sequentially composed of a single patch embedding layer and multiple encoders.  
As shown in Figure~\ref{fig:net5}, each encoder primarily comprises two key components: the Multi-Head Self-Attention (MHSA) and the MLP layer. 
The self-attention calculation process of MHSA can be formulated as follows:
\[
\textbf{X} = \text{Softmax}({\frac{\textbf{Q} \textbf{K}^T}{\sqrt{d_k}}})\textbf{V}, \tag{2}
\]
where \(\textbf{X} \) denotes the tokens updated by the MHSA with a residual, \textbf{Q}, \textbf{K}, and \textbf{V} are respectively the linear transforms of tokens through different project layers, and $d_k$ is the channel dimension of \textbf{K}. 
The updated tokens \(\textbf{X} \) are then sequentially input into a LayerNorm \cite{ln} (\text{LN}) and an MLP layer with a residual, which could be formulated as follows:
\[
\textbf{X}' = \text{MLP}(\text{LN}(\textbf{X})) + \textbf{X}, \tag{3}
\]
where \( \textbf{X}' \) is the final output of the encoder.\\
\textbf{Adapter Fine-Tuning.} Adapter techniques mainly focus on the efficient parameter fine-tuning of Transformer architecture and have achieved advanced performance in computer vision~\cite{adapter1, adapter2,adapter3,adapter4,biadapter,clipadapter}. 
It is commonly integrated in parallel with the MLP layers within each encoder block.
\textit{Different from the full fine-tuning manner that updates all of the model parameters, adapter fine-tuning freezes the abundant pre-trained parameters of the backbone and only updates the scarce parameters within the adapter module.} 
Specifically, an adapter basically consists of a down-projection matrix \( \mathbf{W}_{\text{down}} \in \mathbb{R}^{D \times D^{\prime}} \), a nonlinear mapping \( \mathcal{V} \) (e.g., activation functions or learnable modules), and an up-projection matrix \( \mathbf{W}_{\text{up}} \in \mathbb{R}^{D^{\prime} \times D} \).
The low-rank structure of the adapter (with \( D^{\prime} \ll D \)) ensures that the number of additional trainable parameters is much smaller compared to the entire model.
The vanilla feature pipeline of the adapter can be formulated as follows:
\[
\textbf{X}_{a} = \mathcal{V}(\text{LN}(\textbf{X}) \cdot \mathbf{W}_{\text{down}}) \cdot \mathbf{W}_{\text{up}}, \tag{4}
\]
where \( \textbf{X}_{a} \) denotes the output tokens of the adapter module.
Then \( \textbf{X}_{\text{a}}\) and \( \textbf{X}\) are fused through an influence factor $s$:
\[
\textbf{X}' = \text{MLP}(\text{LN}(\textbf{X})) + s \cdot \textbf{X}_{a} + \textbf{X}. \tag{5}
\]
\begin{figure}[t]
  \centering
   \includegraphics[width=1\linewidth]{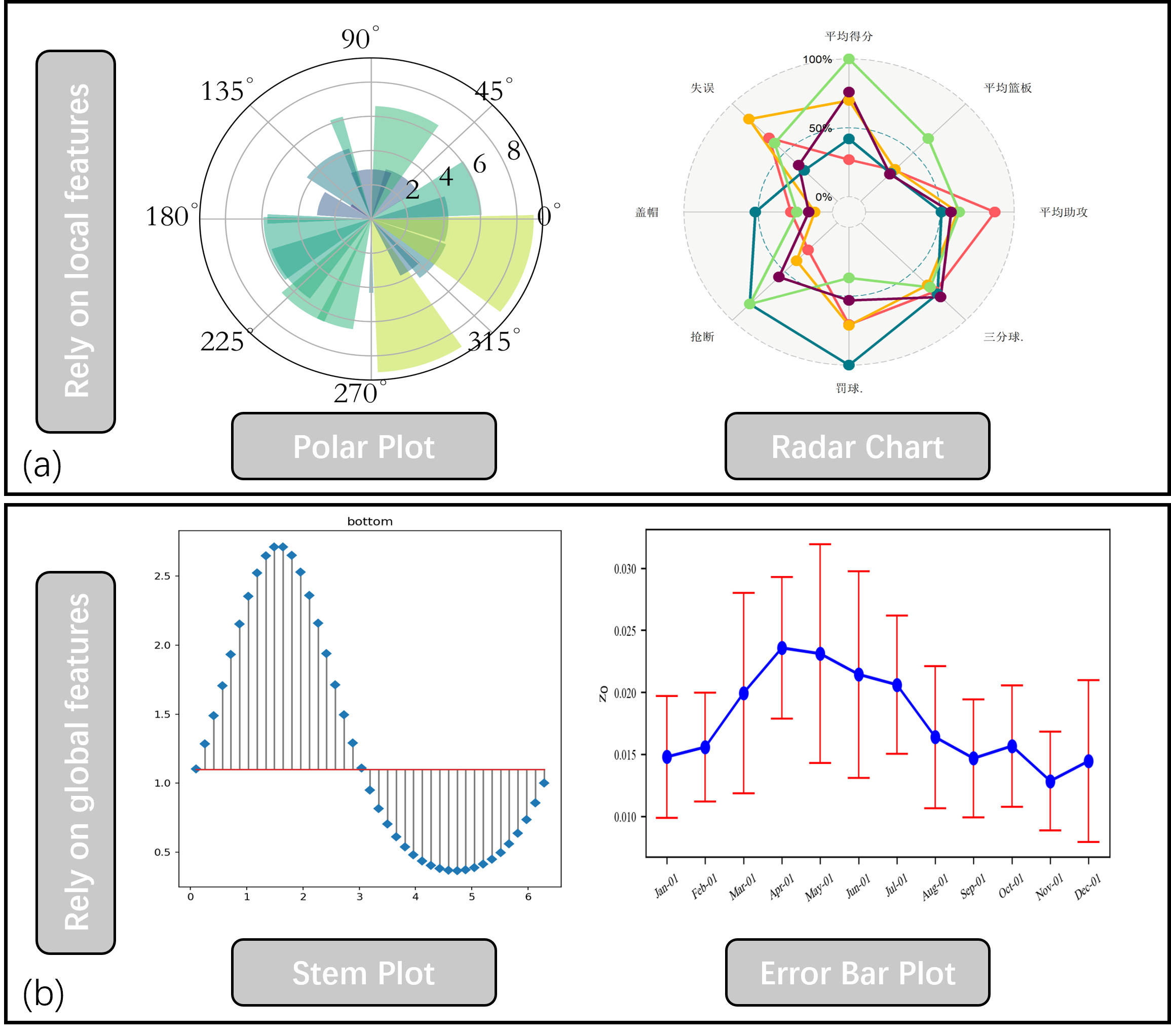}
   \caption{(a) Example of distinction relying on local features. (b) Example of distinction relying on global features.}
   \label{fig:42}
\end{figure}
\subsection{Plot-Adapter}
\subsubsection{Motivation and Overview}
Although significant efforts have been dedicated by previous works to improve the recommendation accuracy in Plot2API~\cite{plot2api,mstil}, there still exists potential challenges arising from multiple scenarios.
Specifically, as different programming languages have unique graphical API sets, the parameters \(\theta\) trained on the dataset of $j$-th language in $\mathcal{G}$ could not be used for the recommendation of other languages.
Similarly, due to the domain gap, the parameters trained on standard plot images are unsuitable for recommendation based on hand-drawn plot images.
These differences necessitate training and storing separate models for each language and domain, leading to significant pressure on parameter growth and computational resource costs.
To address this issue, we propose introducing adapter fine-tuning into Plot2API to only train and store separate adapters instead of the entire model for each language and domain:
\[
\mathbf{I} \overset{F^{\textcolor{blue}{\eta}}_{\textcolor{red}{\phi}_j}(\cdot)}{\rightarrow} \mathbf{Y}, \tag{6}
\]
where $\eta$ and $\phi$ denote the parameters of ViT backbone and adapter, respectively, while the \textcolor{blue}{\textbullet} color indicates \textcolor{blue}{frozen}.

Although the introduction of adapter techniques mitigates the computational burden imposed by the multi-scenario and multi-domain complexity of Plot2API, it also brings specific performance challenges.
In Plot2API, discriminative local features are essential for distinguishing similar plots. 
For example, as shown in Fig.~\ref{fig:42} (a), both within a circular area, a radar plot has multiple axes radiating from the center, while a polar plot uses angles and radii to locate data points.
Similarly, as shown in Fig.~\ref{fig:42} (b), it is easy to distinguish the stem plot from the errorbar plot through local features (stem symbol and vertical line with cap).
However, non-linear transformations \( \mathcal{V} \) in previous adapter techniques are typically composed of simple activation functions (e.g., ReLU, GeLU).
This limits the operations of the entire adapter to the channel level, preventing it from focusing on discriminative regions to boost the frozen backbone.
To address this issue, we propose Plot-Adapter, which integrates a lightweight convolution block within the adapter module for task-specific region capture.

\subsubsection{Network Architecture}
\textcolor{purple}{As shown in Fig.~\ref{fig:net5}, Plot-Adapter adopts a convolutional block to replace the original activation functions as the new nonlinear transformation, which allows the model to focus on discriminative local regions.
Inspired by advanced lightweight CNNs and ViTs~\cite{repvit,lightweightvit1,lightweightvit2,lightweightvit3,lightweightcnn1,lightweightcnn2,lightweightcnn3,mobilevit}, the convolutional block integrated into Plot-Adapter consists of a token mixer, a re-attention layer, and a channel mixer~\cite{metaformer1,metaformer2}.
Specifically, to avoid high computational and parameter costs, we adopt a $3\times3$ deep-wise convolution layer as our token mixer to extract spatial details first.
\[
\textbf{X}_{1} = \mathcal{F}_{Cv}^{3\times3}(\text{LN}(\textbf{X}) \cdot \mathbf{W}_{\text{down}})+\text{LN}(\textbf{X})\cdot \mathbf{W}_{\text{down}}, \tag{7}
\]
where $\mathcal{F}_{Cv}^{i \times i}$ denotes the convolution kernel of $i\times i$ size.
%
%
After that, we utilize a squeeze-and-excitation (SE) module as our re-attention layer to guide the model focusing on more valuable channel signals:
\[
\textbf{X}_{2} = \text{Sigmoid}(\mathcal{F}_{Cv}^{1\times1}(\mathcal{F}_{Cv}^{1\times1}(\mathcal{P}_{Avg}(\textbf{X}_{1})))), \tag{8}
\]
where $\mathcal{P}_{Avg}$ denotes average pooling operation.
Besides, we use two $1\times1$ convolution layers with a GeLU activation function as our channel mixer to mix different channel features and provide non-linear mapping:
\[
\textbf{X}_{a} = (\mathcal{F}_{Cv}^{1\times1}(\text{GeLU}(\mathcal{F}_{Cv}^{1\times1}(\textbf{X}_2)))+\textbf{X}_2)\cdot \mathbf{W}_{\text{up}}. \tag{9}
\]}

\subsubsection{Projection Matrix Sharing}
Although we have tried to lightweight and simplify the CNN blocks, the corresponding introduction will inevitably bring additional model parameters.
To address the parameter growth problem introduced by CNN blocks, inspired by~\cite{ARC}, we introduce a projection matrix sharing strategy to balance model parameters and recommendation performance. 
Specifically, different from most adapter fine-tuning methods that set independent projection matrices for each layer, we set independent CNN blocks for each layer while sharing a projection matrix $\textcolor{green}{\mathbf{W}_{\text{s}}}$.
%
The key insight behind this strategy is that independent projection matrices not only introduce more parameters than lightweight CNN blocks, but also fail to capture spatial information across different layers.
And the channel mixer within the CNN blocks also possesses the feature adjustment capability.
Besides, we apply learnable layer-specific rescaling coefficients $\mathbf{C}$ and biases $\mathbf{B}$ for the sharing projection matrices to preserve flexibility and adaptability, which only introduce a few parameters.
With the projection matrix sharing strategy applied, the feature pipeline of the adapter can be formulated as follows:
\begin{align*}
\textbf{X}_{a} &= \mathcal{V} \left( \text{LN}(\textbf{X})  \cdot \left( \textcolor{green}{\mathbf{W}_{\text{s}}}\mathbf{C}_{\text{down}} + \mathbf{B}_{\text{down}} \right) \right) \\
&\quad \cdot \left( \left( \textcolor{green}{\mathbf{W}_{\text{s}}} \right)^\mathrm{T} \mathbf{C}_{\text{up}} + \mathbf{B}_{\text{up}} \right) \tag{10}
\end{align*}
The color \textcolor{green}{\textbullet} indicates \textcolor{green}{shared} parameters.
In the following experimental section, we have demonstrated that this strategy allows Plot-Adapter to have fewer parameters than the vanilla adapter while maintaining superior performance.

\section{Experiments}
\label{sec:exp}
\begin{table*}[htbp]
\centering
    \resizebox{\textwidth}{!}{
        \begin{tabular}{l|c|ccc|ccc}
            \hline
            \multirow{2}{*}{Method} & \multirow{2}{*}{\parbox{2cm}{\centering Avg. \\ Params(M)}} & \multicolumn{3}{c|}{\textbf{\textcolor{mycolor1}{MAE}}} & \multicolumn{3}{c}{\textbf{\textcolor{mycolor2}{IN21K}}} \\
             & & \textcolor{mycolor1}{Python-13} & \textcolor{mycolor1}{R-32} & \textcolor{mycolor1}{HDpy-13} & \textcolor{mycolor2}{Python-13} & \textcolor{mycolor2}{R-32} & \textcolor{mycolor2}{HDpy-13} \\
            \hline
            \rowcolor{gray!20}
            Full-tuning & 85.81 & 64.49 & 40.73 & 78.73 & 67.65 & 43.73 & 80.12 \\
            Linear & 85.81 & 62.78 & 40.78 & 76.65 & 66.90 & 40.34 & 79.23 \\
            \hline
            \hline
            VPT-4 & 0.04 & 60.14 & 38.67 & 76.81 & 66.84 & 42.53 & 76.18 \\
            VPT-8 & 0.07 & 61.53 & 39.13 & 74.00 & 66.95 & 42.76 & 75.90 \\
            VPT-16 & 0.15 & 61.16 & 39.16 & 75.24 & 65.85 & 41.53 & 79.44 \\
            \hline
            \hline
            AdaptFormer-4 & 0.08 & 61.40 & 38.85 & 75.38 & 67.02 & 42.41 & 79.09 \\
            AdaptFormer-16 & 0.30 & 62.05 & 39.83 & 75.32 & 66.62 & 41.78 & 76.67 \\
            AdaptFormer-64 & 1.19 & 61.13 & 39.41 & 73.22 & 66.71 & 42.67 & 78.59 \\
            \hline
            \hline
            Plot-Adapter-V1-4 & 0.12 & 68.90 & 40.50 & 77.62 & 68.16 & 44.47 & 81.36 \\
            Plot-Adapter-V1-16 & 0.35 & 67.02 & 40.30 & 78.47 & 68.29 & 43.36 & 81.07 \\
            Plot-Adapter-V1-64 & 1.31 & 67.33 & 40.78 & 81.22 & 67.86 & 43.38 & 83.05 \\
            \hline
            \hline
            Plot-Adapter-V2-4 & 0.06 & 69.10 & 40.89 & 79.24 & 67.26 & 43.62 & 82.05 \\
            Plot-Adapter-V2-16 & 0.08 & 69.12 & 40.35 & 79.09 & 68.16 & 44.30 & 81.89 \\
            Plot-Adapter-V2-64 & 0.19 & 69.09 & 40.92 & 79.90 & 68.78 & 44.32 & 83.40 \\
            \hline
        \end{tabular}}
        \caption{Results of different methods and settings on three downstream datasets, in which V1 and V2 denote the versions of Plot-Adapter without and using the projection matrix sharing strategy, respectively.  }
        \label{table1}

\end{table*}
\begin{table}[h]
    \centering
    \label{tab:experiment_results}
    \renewcommand{\arraystretch}{1.7}
    \resizebox{\linewidth}{!}{
        \begin{tabular}{c| c| c| c|c}
            \hline
            \textbf{Token Mixer} & \textbf{Attention} & \textbf{Channel Mixer} & \textbf{mAP}&Params \\
            \hline            
            $\mathcal{F}_{Cv}^{3\times3}$+$\mathcal{F}_{Cv}^{5\times5}$+$\mathcal{F}_{Cv}^{7\times7}$ & \cellcolor{green!20}SE~\cite{se} & $D^{\prime} \to D^{\prime}$ & 79.44 & 0.25\\
            $\mathcal{F}_{Cv}^{3\times3}$+$\mathcal{F}_{Cv}^{5\times5}$+$\mathcal{F}_{Cv}^{7\times7}$ & SA~\cite{sa} & $D^{\prime} \to D^{\prime}$ & 77.93 & 0.22\\
            \hline                      
            $\mathcal{F}_{Cv}^{3\times3}$+$\mathcal{F}_{Cv}^{5\times5}$+$\mathcal{F}_{Cv}^{7\times7}$ & SE~\cite{se} & $D^{\prime} \to 2D^{\prime} \to D^{\prime}$ & 77.94 & 0.40\\            $\mathcal{F}_{Cv}^{3\times3}$+$\mathcal{F}_{Cv}^{5\times5}$+$\mathcal{F}_{Cv}^{7\times7}$ & SE~\cite{se} & \cellcolor{green!20}$D^{\prime} \to\frac{D^{\prime}}{2} \to D^{\prime}$ & 81.04 & 0.25\\
            \hline            
            \cellcolor{green!20} $\mathcal{F}_{Cv}^{3\times3}$ & SE~\cite{se} & $D^{\prime} \to\frac{D^{\prime}}{2} \to D^{\prime}$ & \textbf{81.22} & 0.19\\
            $\mathcal{F}_{Cv}^{3\times3}$ $\cdot$  $\mathcal{F}_{Cv}^{3\times3}$ & SE~\cite{se} & $D^{\prime} \to\frac{D^{\prime}}{2} \to D^{\prime}$ & 79.28 & 0.20\\
            \hline
        \end{tabular}
    }
    \caption{Ablation experiments of lightweight CNN block architecture in Plot-Adapter on HDpy-13.}
    \label{table2}
\end{table}

\subsection{Datasets Overview}
We conducted extensive experiments on Python-13, R-32, and HDpy-13 datasets to validate the effectiveness of Plot-Adapter \cite{plot2api}. 
In addition to HDpy-13, we describe the former two datasets as follows:\\
\textbf{Python-13.} This dataset consists of 6,350 images generated through APIs in the Python programming language (5,080 for training and 1,270 for test). 
It is a collection of posts about the Python programming language from the Stack Overflow Data Dump and covers 13 categories of graphical APIs in the Python programming language.\\
\textbf{R-32.} This dataset includes 9,114 images generated through APIs in the R programming language (7,292 for training and 1,822 for test). 
It is similarly taken from the Stack Overflow Data Dump and encompasses 32 categories of graphical APIs in the R programming language.\\
%
%
%
\subsection{Evaluation Metrics}
We use Average Precision (AP) as the metric to evaluate the recommendation accuracy for each API.
The AP is computed as the area under the precision-recall (P-R) curve. 
To assess the overall model performance on the target scene, we report the mean value of the APs over all APIs known as the mean average precision (mAP).
%
\subsection{Experimental Setup}
The Vision Transformer (ViT-B/16) \cite{vanillavit} is used as our backbone for all experiments.
Besides, we employ ImageNet-21k~\cite{imagenet} supervised pre-trained model and MAE~\cite{mae} self-supervised model.
Our experiments are based on Pytorch\cite{pytorch} framework running on two NVIDIA A800 GPUs.
All models are trained for 100 epochs with a batch size of 32, while Adam~\cite{adam} is employed for optimization.
The learning rate is adjusted using cosine annealing with an initial value of $1\times10^{-5}$ to improve convergence.
We employ the MultiLabelSoftMarginLoss, which is designed for scenarios where labels are probabilistically independent but can jointly occur.
The images will be resized to 224 × 224 and then normalized before input into models, while random horizontal flipping and random erasing are applied only during training for further data augmentation. 

\subsection{Comparison methods.} 
We compare our proposed Plot-Adapter with four different fine-tuning methods:
(1) Full Fine-tuning: fine-tuning all the parameters in the model.
(2) Linear Probing: fine-tuning only a linear layer used for prediction.
(3) Visual Prompt Tuning (VPT)~\cite{vpt}: fine-tuning only the extra token parameters and a linear layer.
(4) AdaptFormer~\cite{adaptformer}: fine-tuning only the extra vanilla adapter parameters and a linear layer.

\subsection{Quantitative Analysis}
The main experiment results on three datasets are reported in Tab.~\ref{table1}.
The strength of our method will be discussed in the following three basic dimensions, \ie, recommendation accuracy, number of fine-tuning parameters, and generalization ability.
Firstly, it can be observed that compared to the vanilla adapter and other comparison methods, Plot-Adapter has achieved the most advanced recommendation accuracy, which highlights the involvement of convolution operations in the adapter.
However, these performance improvements have not significantly increased the number of additional fine-tuning parameters, which benefitted from the applied lightweight CNN block.
In addition, the reduced parameters in the projection matrix enable Plot-Adapter to have even fewer parameters than the vanilla adapter while still maintaining competitive performance.
Besides, the unanimous improvements under different programming languages, scenarios, pre-trained weights, and hyperparameter settings demonstrate the generalization ability of our proposed Plot-Adapter.

\subsection{Ablation Study}
We first conduct ablation experiments to evaluate the effectiveness of lightweight CNN block architecture in Plot-Adapter, which are shown in Tab.~\ref{table2}.
It could be observed that the re-attention at the channel level is more beneficial than that at the spatial level, which is because the convolutional layers in Plot-Adapter of each transformer layer are already capable of progressively focusing on local discriminative features.
Moreover, we are not surprised to see that deep channel mixers bring more accuracy gains, but excited to find that squeezing channel dimensions leads to better performance than expanding, as the former requires fewer parameters.
Besides, while many works have demonstrated the importance of combining multi-scale features~\cite{kiprn,ncmnet,pyconv,qin2023asdfl,dappm,qin2025boosting}, we find that in Plot-Adapter, a single 3×3 convolution layer is sufficient to serve as a satisfactory token mixer with least parameters.
As shown in Tab.~\ref{table3}, we further conduct ablation experiments to evaluate the effect of different parameter initialization manners in projection matrix sharing, which point out that uniform initialization is better than zero one.

\begin{table}[h]
    \centering
    \label{tab:experiment_results}
    \renewcommand{\arraystretch}{1}
    \resizebox{\linewidth}{!}{
        \begin{tabular}{l l l l c}
            \toprule
            down\_rescale & down\_bias & up\_rescale & up\_bias & mAP \\
            \midrule
            uniform  & zeros  & zeros  & zeros  & 79.41 \\
            zeros  & zeros  & uniform  & zeros  & 78.61 \\
            \rowcolor{green!20} uniform  & zeros  & uniform  & zeros  & 79.90 \\
            zeros  & zeros  & zeros  & zeros  & 75.44 \\
            \bottomrule
        \end{tabular}
    }
    \caption{Ablation experiments of parameter initialization manners in projection matrix sharing on HDpy-13.}
    \label{table3}
\end{table}
\section{Conclusion}
To provide a simple and quick data visualization solution for non-experts and beginners, we propose a hand-drawn plot dataset called HDpy-13 to improve the graphical API recommendation performance for hand-drawn plot images.
In addition, to reduce the significant pressure of parameter growth and computational resource costs caused by multi-domain and multi-language in Plot2API, we propose Plot-Adapter to train and store separate adapters instead of an entire model for each language and domain. 
Specifically, Plot-Adapter integrates a lightweight CNN block to enhance the local feature capture ability and introduces projection matrix sharing to further compress the fine-tuning parameters.
Experimental results demonstrate both the effectiveness of HDpy-13 and the efficiency of Plot-Adapter.
{
    \small
    \bibliographystyle{ieeenat_fullname}
    \bibliography{main}
}
\end{document}

%% file: preamble.tex
\usepackage[dvipsnames]{xcolor}
